\documentclass[sigconf]{acmart}
\usepackage{listings}
\usepackage{xcolor}
\usepackage{graphicx}
\usepackage{amsmath}
\usepackage{listings}
\usepackage{float}
\usepackage{floatflt}
\usepackage{comment}
\usepackage{graphicx}
\usepackage{textcomp}
\usepackage{hyperref}
\usepackage{makecell}

\renewcommand\footnotetextcopyrightpermission[1]{} 
\begin{document}
\title{Improving Artifact Robustness for CT Deep Learning Models Without Labeled Artifact Images via Domain Adaptation}

\author{Justin Cheung}
\email{jcheun11@jh.edu}
\orcid{}
\affiliation{%
  \institution{Johns Hopkins University}
  \city{Baltimore}
  \state{Maryland}
  \country{USA}
  }

\author{Samuel Savine}
\email{ssavine1@jh.edu}
\orcid{}
\affiliation{%
  \institution{Johns Hopkins University}
  \city{Baltimore}
  \state{Maryland}
  \country{USA}
  }
  
\author{Calvin Nguyen}
\email{cnguye89@jh.edu}
\orcid{}
\affiliation{%
  \institution{Johns Hopkins University}
  \city{Baltimore}
  \state{Maryland}
  \country{USA}
  }
  
\author{Lin Lu}
\email{llu45@jhu.edu}
\orcid{}
\affiliation{%
  \institution{Johns Hopkins University}
  \city{Baltimore}
  \state{Maryland}
  \country{USA}
  }

\author{Alhassan S. Yasin}
\email{ayasin1@jhu.edu}
\orcid{0009-0001-8033-9850}
\affiliation{%
  \institution{Johns Hopkins University}
  \city{Baltimore}
  \state{Maryland}
  \country{USA}
}

\begin{abstract} 
Deep learning models which perform well on images from their training distribution can degrade substantially when applied to new distributions. If a CT scanner introduces a new artifact not present in the training labels, the model may misclassify the images. Although modern CT scanners include design features which mitigate these artifacts, unanticipated or difficult-to-mitigate artifacts can still appear in practice. The direct solution of labeling images from this new distribution can be costly. As a more accessible alternative, this study evaluates domain adaptation as an approach for training models that maintain classification performance despite new artifacts, even without corresponding labels. We simulate ring artifacts from detector gain error in sinogram space and evaluate domain adversarial neural networks (DANN) against baseline and augmentation-based approaches on the OrganAMNIST abdominal CT dataset. We simulate the absence of labels from an unseen distribution via masking in the loss function and selectively detaching unlabeled instances from the computational graph. Our results demonstrate that baseline models trained only on clean images fail to generalize to images with ring artifacts, and traditional augmentation with other distortion types provides no improvement on unseen artifact domains. In contrast, the DANN approach improves classification accuracy on ring artifact images using only unlabeled artifact data during training, demonstrating the viability of domain adaptation for artifact robustness. The domain-adapted model achieved a classification accuracy of 77.4\% on ring artifact test data, 38.7\% higher than a baseline model only trained on images with no artifact. These findings provide empirical evidence that domain adaptation can effectively address distribution shift in medical imaging without requiring expensive expert labeling of new artifact distributions, suggesting promise for deployment in clinical settings where novel artifacts may emerge.

\end{abstract}


\maketitle
\fancyfoot{}
\fancyhead{}
\thispagestyle{empty}

\section{Introduction}
\label{sec:introduction}
Deep learning models have shown strong performance on image classification tasks when the training and testing data are derived from similar distributions. However, even small shifts in data distribution, such as changes in resolution or noise, can cause significant drops in model performance compared to human performance \cite{dodge2017study}. This performance degradation due to \textit{domain shift} is important to consider in medical imaging, where model predictions may directly impact clinical decisions.

Medical images seen by a model at inference time can exhibit a domain shift compared to those seen by the model during training due to differences in hardware, patient populations, and acquisition protocols, making robustness to common domain shifts critical for safe deployment. Although augmentation of training data to represent a wider variety of images is known to be an effective method to combat these differences in image processing, a mismatch between the distributions of the augmented dataset and the dataset seen at inference time can still lead to poor performance on the latter \cite{geirhos2018generalisation}. It is possible for a new type of domain shift to be seen in the field which was not accounted for during augmentation - for instance, a method for augmenting the dataset to accurately depict the domain shift may not be known, or we may seek to apply the model to images produced by a new scanner or to a new patient population. Given the high cost and demand for expert physicians' time required for labeling medical images for tasks of medical interest \cite{rahimi2021addressing}, requiring labeling of images from new distributions to improve supervised model performance may not be a practical or desirable solution.

Computed tomography (CT) in particular is known to suffer from artifacts such as motion blur, beam hardening, and metal-induced streaks. These artifacts degrade image quality and complicate diagnosis. These artifacts often necessitate repeat scans, increasing both healthcare costs and patient exposure to ionizing radiation \cite{barrett2004artifacts}. In the event that these are not represented in the training set, we can expect poor classification performance.

Our goal is to create a system capable of accurately diagnosing underlying conditions in distorted CT scans (those with the aforementioned artifacts), thus minimizing the need for repeat imaging. This work is focused exclusively on CT images and a select set of domain shifts. Specifically, the objective is to develop a deep image classifier that maintains robust classification performance even in the presence of artifacts unrepresented in the labeled training images, reducing radiation exposure by limiting unnecessary scans.

To reach this goal, this study will train a deep image classifier that can generalize across artifact domains by leveraging \textit{domain adaptation} to achieve high classification accuracy without requiring labeled images exhibiting the new domain shift. We use a physics-informed artifact simulation technique to characterize the CT-relevant ring artifact, as a medically relevant proof of concept to allow fair comparison and assessment of generalization techniques. The methodology described here has potential to be applied when images of a new domain shift appear in the field - the new images can be used to aid in supervised training even without their labels via domain adaptation.

\subsection{Related Work}

Geirhos et al. conducted a study comparing human classification ability to that of deep learning models in the presence of distortions, which we can consider as domain shifts. They trained ResNet-50 on a 16-class variant of the ImageNet dataset, and evaluated the model on distorted versions of images it had not seen during training. They concluded that data augmentation is insufficient for preparing models for unseen distortions \cite{geirhos2018generalisation}.

In response to these findings, we consider domain adaptation as a potential improvement over data augmentation in the scenario that images from the unseen distortion are available, but without labels. Normally, models only trained on data from one distribution (a \textit{source domain}) cannot generalize to data from a new distribution (a \textit{target domain}), even if the two domains are related. Domain adaptation techniques allow the model to see the target domain without labels during training. Unlike conventional supervised learning, domain adaptation techniques allow unlabeled data to guide supervised training \cite{liu2022deep}. Domain adaptation techniques have previously been applied for CT, such as to adapt a metal artifact reduction network \cite{du2023deep} as well as to CT/MRI cross-modality transfer learning \cite{chen2020unsupervised, zhao2022uda}; in this work, we explore the applicability of domain adaptation for artifact robustness of deep learning models.

One such domain adaptation approach is the domain adversarial neural network (DANN), which allows a model to learn domain-invariant features via a \textit{domain classifier} with a \textit{gradient reversal layer} during classification training \cite{ajakan2014domain, ganin2016domain}. This intuitive approach discourages the learning of features which help with domain classification, instead forcing the model to learn features which are not specific to a domain. For our study, we elected to use this approach due to its relatively simple implementation for our first steps of exploring domain adaptation techniques.

\section{Methods}

\subsection{Dataset}

We identified MedMNIST as a valuable source of labeled images for a variety of medical imaging modalities and classification tasks. We chose MedMNIST's OrganAMNIST abdominal organ classification dataset in particular since its images come from axial CT \cite{yang2023medmnist}, which the sinogram manipulation technique we used directly applies to.

We use the existing set of 11 possible organ labels: bladder (0), femur-left (1), femur-right (2), heart (3), kidney-left (4), kidney-right (5), liver (6), lung-left (7), lung-right (8), pancreas (9), and spleen (10).

For cross validation experiments, we split the dataset's provided 34,561 training samples and 6,491 validation samples into 5 folds, and test on the provided 17,778 test samples.

We determined that the following 4 distortions were relevant to CT imaging, and thus applied them to OrganAMNIST:
\begin{itemize}
\item No distortion
\item Uniform noise (within +/- 35\% of image intensity range)
\item Rotate 90°
\item Ring artifact
\end{itemize}

\begin{figure}[ht!]
  \includegraphics[width=\linewidth]{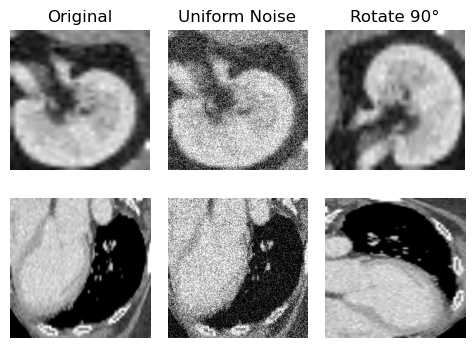}
  \caption{Samples from our distorted OrganAMNIST dataset exhibiting no distortion, uniform noise, and rotation by 90°.}
  \label{figOrganAMNISTDistortionExamples}
\end{figure}

For each of the 4 distortions, a copy was made of the training, validation, and test sets with the distortion applied. Examples for the first few distortions are depicted in Fig.~\ref{figOrganAMNISTDistortionExamples}.

\subsection{Synthetic CT Distortion}

We emulate distortions specific to CT based in the physical principles of CT image acquisition via an approach based on the Radon transform.

To situate our work in CT imaging fundamentals - recall that in the simplest parallel beam scheme, the \textit{forward projection} is a discrete approximation of the Radon transform which takes 1D projections of a 2D axial slice of a volume such as a patient. These projections are taken at evenly spaced angles about the scanner's central axis (isocenter), and are stacked as a 2D sinogram. For a scanner, the projections are collected by transmitting X-rays along slices through the volume; in our case, we treat a CT scanner's output image itself as an approximation of a slice through the original volume, and project through this image. CT reconstruction is then performed via \textit{backprojection}, a discrete approximation of the inverse Radon transform which is computed both in practice and in our simulated scheme.

At a high level, our approach is to:
\begin{enumerate}
\item Perform forward projection on an image to mimic the transmission of X-rays through the body, resulting in an emulated sinogram
\item Apply a distortion to this emulated sinogram in a manner consistent with undesirable scanner imperfections or physical phenomena related to the absorption, scattering, detection, etc. of X-rays during CT scanning
\item Perform backprojection on the distorted sinogram to reconstruct a distorted image
\end{enumerate}

We use a number of projection angles equal to the original width of the image in pixels, as advised by \cite{scikit-image}. For proof of concept, we attenuate each row of the sinogram to emulate gain error in each X-ray detector, resulting in the ``ring artifact" distortion. A similar approach was used by \cite{an2020ring} to test their ring artifact reduction algorithm.

For our experiments, our distortion function applies a random multiplicative attenuation to all simulated CT detectors (rows of the sinogram). This is based on the known phenomenon where minor gain error in CT scanner X-ray detectors causes visible rings to appear in the final reconstructed CT image, “typically within a few percent” \cite{blaj2019dead}. The gain error is uniformly distributed in $[-10\%, +10\%]$ following the range used by \cite{an2020ring} (in our initial experiments, attenuation up to 3\% was not challenging for the models to adapt to when provided with labeled original data during training).

Unfortunately, since our dataset lacks metadata specifying patient position relative to the scanner, there is not much basis to decide a reasonable translation for the ring artifacts. This is an unfortunate limitation, as rings often are translated relative to patient anatomy since the rings are centered on the scanner isocenter, while patient anatomy and the field of view used for the final image can each be shifted relative to the isocenter. To try to mimic minor differences among patients' positioning, we arbitrarily shift the scanner center by up to 10 pixels horizontally and up to 10 pixels vertically. Because the forward projection implementation treats the image center as the isocenter, we translate the rings by carefully zero-padding the image prior to applying the forward projection so that the isocenter is the padded image's center. In particular, to implement a signed shift horizontally by $dx$ and vertically by $dy$ to a square image of height $H$, a pad of $\max(|dx|,|dy|)+dy$ is applied to the image top and $\max(|dx|,|dy|)+dx$ is applied to the image left, while maintaining an overall image width of $M=2(\max(|dx|,|dy|)+\frac{H}{2})$ in either direction. The zero-padded region makes no contribution to the projections, and the resulting rings are shifted. To avoid losing nonzero content of the image during forward projection as the image is effectively rotated for each projection, the image is further padded to at least the width of its diagonal, $\left\lceil M\sqrt{2}\right\rceil$. 

\paragraph{Proposed Algorithm for Reconstruction Error Mitigation}

\begin{figure}[ht!]
  \includegraphics[width=\linewidth]{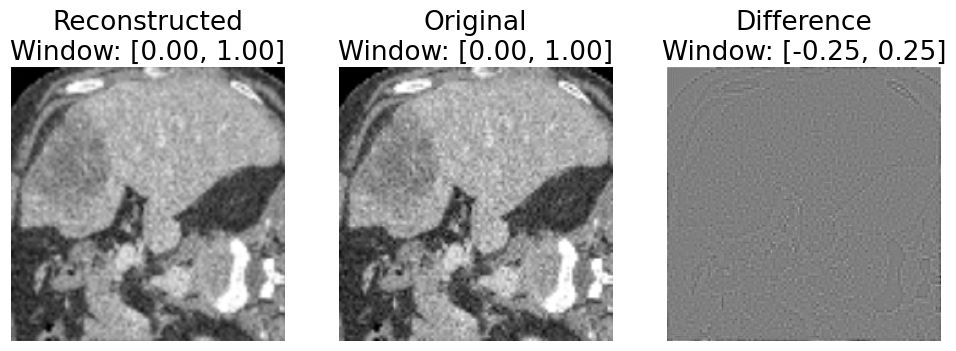}
  \caption{Reconstruction error example (leftmost image is $R$, middle image is $O$). The rightmost figure is the difference $R-O$; there is visible non-uniform noise roughly following anatomical structures which is visible between the original image $O$ and the result after forward projection and backprojection $R$. In all figures of CT images, the \textit{window} is expressed in the format $[a,b]$ where pixel intensities below $a$ are depicted in black, and pixel intensities above $b$ are depicted in white.}
  \label{figReconError}
\end{figure}

\begin{figure}[ht!]
  \includegraphics[width=\linewidth]{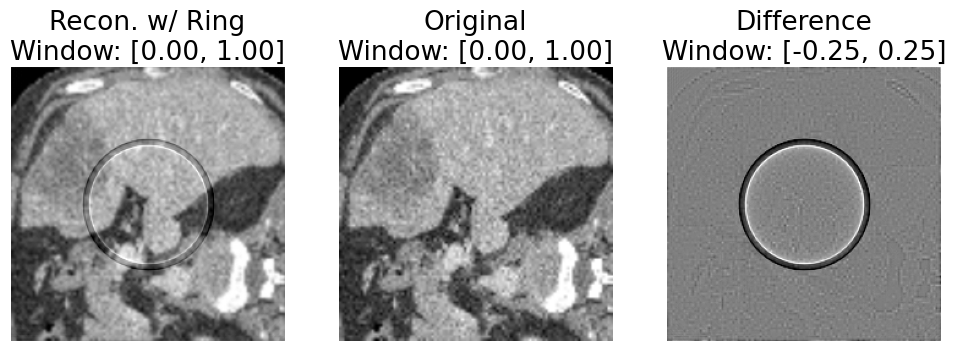}
  \caption{Unintended reconstruction error mixed with intended simulated ring artifact (leftmost image is $R_{distorted}$, middle image is $O$, rightmost image is $R_{distorted}-O$). The error introduced by forward and backprojection makes it difficult to claim that $R_{distorted}$ represents the ring artifact in isolation, as $R_{distorted}$ can also be distinguished from the original image based on the error. For ease of visualization, we simulate 5 adjacent detector channels with gain error -10\%.}
  \label{figReconErrorAndRing}
\end{figure}

We observe a notable “reconstruction error” introduced by our simulation approach. In particular, when considering discrete images, the result of performing forward projection followed by backprojection is not equivalent to the original image, as seen in Fig.~\ref{figReconError}. We were not able to mitigate this error by increasing the number of projection samples to increase the sampling resolution for the transform. We believe this may be a consequence of using relatively small images with far fewer samples than can be obtained with a real CT scanner and a physical volume. Unfortunately, with our approach, this unintended reconstruction error becomes mixed with the intentional simulated distortions we apply in sinogram space (Fig.~\ref{figReconErrorAndRing}), muddying the conclusions which can be drawn regarding domain adaptation.

To reduce reconstruction error in the distorted image in our simulation scheme, we propose the following algorithm to produce the distorted image.

Consider a space of images (of a certain size) $\mathcal{I}$, and an original image before distortion $O\in\mathcal{I}$. Denote an implementation of the forward transform as $\text{radon}:\mathcal{I}\rightarrow\mathcal{V}$ where $\mathcal{V}$ represents a space of sinograms  (alternatively known as view space). Denote an implementation of the corresponding backprojection as $\text{iradon}:\mathcal{V}\rightarrow\mathcal{I}$ - we use the scikit-image implementation. Additionally consider a sinogram-space distortion function $\text{distort}:\mathcal{V}\rightarrow\mathcal{V}$. The algorithm is simply the calculation of $O_{distorted}$ via the steps in \ref{eqCleanReconError}.

\begin{equation}
\begin{aligned}
S &= \text{radon}(O), &
S_{distorted} &= \text{distort}(S), \\
R &= \text{iradon}(S), &
R_{distorted} &= \text{iradon}(S_{distorted}), \\
D &= R_{distorted} - R, &
O_{distorted} &= O + D.
\end{aligned}
\label{eqCleanReconError}
\end{equation}

The underlying assumption of this approach is that a similar reconstruction error is carried by both $R_{distorted}$ and $R$, so their difference image should be an additive representation of the intended distortion applied in sinogram space with decreased reconstruction error. We then assume that adding this cleaner version of the distortion directly to the original image produces a distorted image with a mitigation of the unintended reconstruction error.

\begin{figure}[ht!]
  \includegraphics[width=\linewidth]{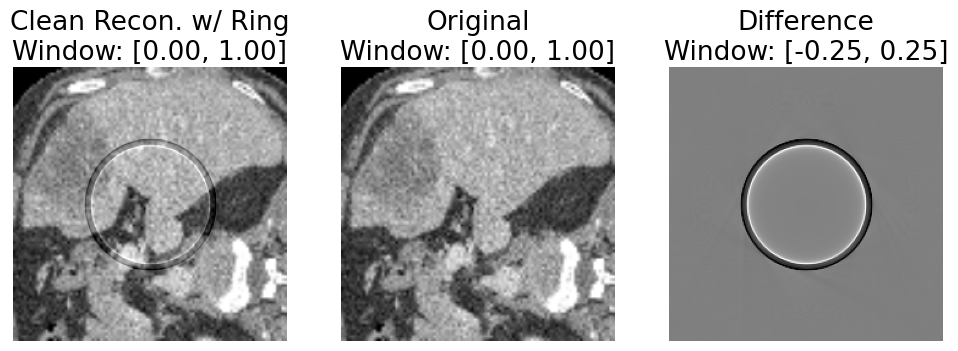}
  \caption{Result of our proposed approach to addressing error introduced by simulating artifacts with forward and backprojection. Leftmost image is $O_{distorted}$, middle is $O$, right is $O_{distorted}-O$. There is no longer a visible difference between the distorted image $O_{distorted}$ and $O$ aside from the ring artifact; this makes $O_{distorted}$ a reasonable depiction of $O$ with the ring artifact, and without error introduced by forward and backprojection. For ease of visualization, we simulate 5 adjacent detector channels with gain error -10\%.}
  \label{figCleanRing}
\end{figure}

\begin{figure}[ht!]
  \includegraphics[width=\linewidth]{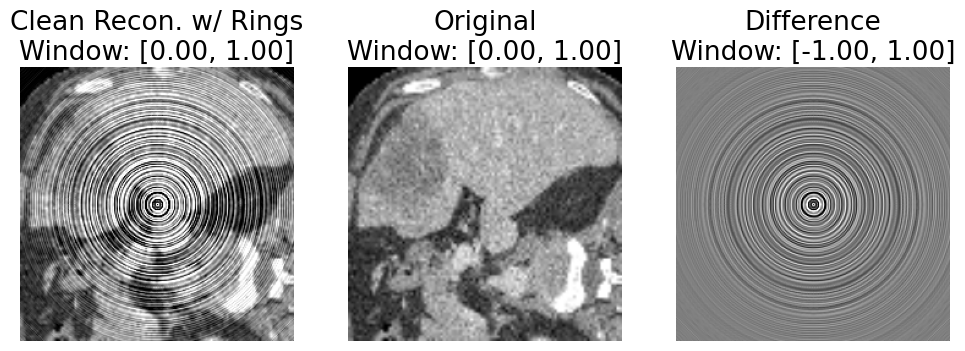}
  \caption{A sample from our distorted OrganAMNIST dataset exhibiting our simulated ring artifact with up to 10\% gain error in all simulated detectors.}
  \label{figRingExample}
\end{figure}

See Fig.~\ref{figCleanRing} for an example following those used in discussion thus far, in which 10\% gain error is simulated on 5 adjacent detector channels to make the ring easier to distinguish from anatomy for discussion's sake. See Fig.~\ref{figRingExample} for an example of the actual ring distortion we used in our experiments, which applies up to 10\% gain error to all simulated detectors.

\subsection{Setup, Training and Evaluation}

\subsubsection{Architecture} 

Recall that in a domain adaptation setting, images come from either the source domain $\mathcal{S}$ or the target domain $\mathcal{T}$, each of which in this context are subsets of $\mathcal{I}$. $\mathcal{S}$ and $\mathcal{T}$ are related but distinct distributions, separated by some domain shift. Images in $\mathcal{S}$ have labels, while images in $\mathcal{T}$ lack labels; our goal is to use $\mathcal{S}$ to learn the supervised classification task, while using both $\mathcal{S}$ and $\mathcal{T}$ to adapt the model to classify well on $\mathcal{T}$ despite the absence of its labels. A domain label $d$ will be used to indicate an instance's membership in either $\mathcal{S}$, for which $d=0$; or $\mathcal{T}$, for which $d=1$. \cite{ganin2015unsupervised}

\begin{figure}[ht!]
  \includegraphics[width=0.9\linewidth]{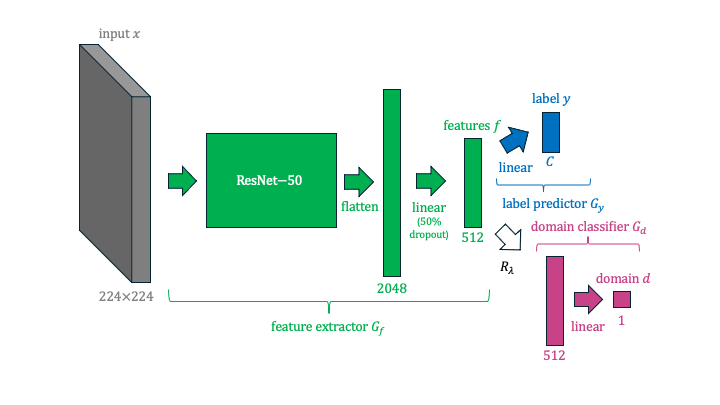}
  \caption{DANN using ResNet-50 as a feature extractor, used in our domain adaptation experiments.}
  \label{figModelArchitecture}
\end{figure}

Our model architecture, depicted in Fig.~\ref{figModelArchitecture}, is adapted from the technique introduced by \cite{ganin2015unsupervised} to achieve this end, in which a domain classifier is added to an existing classification CNN, and its gradients are negated and scaled before being backpropagated further into the lower layers of the network. 

Specifically, \cite{ganin2015unsupervised} breaks their architecture into three networks: a feature extractor $G_f: \mathcal{I}\rightarrow\mathcal{F}$ where $\mathcal{F}$ is an arbitrary feature space, a label predictor $G_y: \mathcal{F}\rightarrow\mathbb{R}^{C}$ which maps latent features to probabilities for each of the valid labels in the classification problem ($C=11$ in the case of OrganAMNIST prediction), and a domain classifier $G_d:\mathcal{F}\rightarrow\mathbb{R}$ which maps latent features to a probability of the instance being from the target domain. A gradient reversal layer $R_\lambda$ is also included, whose sole hyperparameter is adaptation rate $\lambda$; during forward propagation, $R_\lambda$ acts as the identify transform ($R_\lambda(\mathbf{x})=x$), while during backpropagation, $R_\lambda$ scales the backpropagated gradient by $-\lambda$ ($\frac{dR_\lambda}{d\mathbf{x}}=-\lambda I$ where $I$ is the identity matrix, as per \cite{ganin2015unsupervised}). Intuitively, the gradient reversal layer moves feature extractor parameters in the opposite direction of what would help the domain classifier's performance. This forces features learned by the feature extractor to simultaneously contain as little useful information for the domain classifier to predict domain from as possible, while also still containing information which allows the label predictor to correctly classify on the labeled source dataset.

Given an instance $\mathbf{x}\in \mathcal{I}$, our full DANN can be expressed in the usual formulation as feature extraction $\mathbf{f}=G_f(\mathbf{x})$, label prediction $\mathbf{\hat y}=G_y(\mathbf{f})$, and domain classification $\hat d = G_d(R_\lambda(\mathbf{f}))$. In our implementation, $G_f$ starts with ResNet-50 \cite{he2016deep} outputting a flat vector of features of width 2048. We follow this by a linear layer of width 512 with ReLU activation. A 50\% dropout layer is used at this layer to help avoid overfitting. $G_y$ is a single linear layer of width $C$ followed by softmax. $G_d$ is a single linear layer of width $d$ followed by sigmoid.

\subsubsection{Loss Function}

We implement an equivalent loss function to the one described in \cite{ganin2015unsupervised}.

To establish notation - given a matrix $A$, denote the element of $A$ at the $i$-th row, $j$-th column as $a_{i,j}$. Also, denote the $i$-th row of $A$ as $a_{i,:}$. Likewise, given a vector $\mathbf{v}$, its $i$-th element is denoted $v_i$.

Consider a minibatch of $N$ elements, whose true labels are $\mathbf{y}\in\{1..C\}^{N}$ and whose true domain classes are $\mathbf{d}\in\{0,1\}^N$. Consider the natural generalization of our model to minibatches to produce predictions $\hat Y \in \mathbb{R}^{N\times C}$, $\mathbf{\hat d} \in \mathbb{R}^N$. 

The loss function from \cite{ganin2015unsupervised, ganin2016domain} over a minibatch can be expressed as \eqref{eqLoss}:

\begin{multline}
\mathcal{L}(\hat Y, \mathbf{y}, \mathbf{\hat d}, \mathbf{d}) =\\ \sum_{i\in\{1..N\}}(1-d_i)\mathcal{L}_{CE}(\hat y_{i,:}, y_i) + \sum_{i\in\{1..N\}}\mathcal{L}_{BCE}(\hat d_i, d_i) 
\label{eqLoss}
\end{multline}

Note that the label predictor $G_y$ uses cross entropy loss $\mathcal{L}_{CE}$, while the domain classifier $G_d$ uses binary cross entropy loss $\mathcal{L}_{BCE}$.

Since we simulate artifacts for instances of the OrganAMNIST dataset, which means we have the organ labels for the target domains in our experiments, we simulate the effect of having no labels by masking out the effect of the target domain's labels in our implementation's loss function calculation. Here, we use $(1-d_i)$ to mask out contributions of target domain instances to the loss function of the label predictor. We furthermore detach the logits of target domain samples produced by the label predictor from the computational graph. This simplifies our implementation of the architecture in the PyTorch framework so that we can perform stochastic gradient descent over minibatches without special handling to prevent target domain instances from propagating forward through the label predictor.

\subsubsection{Experimental Design} 

For a fair comparison between augmentation and domain adaptation via DANN, we split our study into four experiments.

\paragraph{Experiment 1} This experiment allows us to check that the same baseline lack of generalization seen with ImageNet by Geirhos et al. is also present in the new OrganAMNIST dataset we are interested in. This helps us demonstrate that the generalization task is not trivial for this dataset. We hypothesize that like in \cite{geirhos2018generalisation}, the model will perform well on the distortion type it has seen during training. We expect that the model will not be able to generalize to distortions it has not seen before.

We use the OrganAMNIST dataset with the corresponding set of 4 distortions mentioned earlier. For each of the 4 distortions, a copy of each of the training, validation, and test sets is created where the distortion has been applied to all samples. For each of the 4 distortions, we train an instance of ResNet-50 only with the training samples with that distortion applied. This results in 4 models. For each of the 4 models, we test the model on all 4 distortions’ test data.
We use stochastic gradient descent with a learning rate of 0.1, a linear decay learning rate schedule, and a weight decay of $10^{-4}$. We train for 50 epochs.

\paragraph{Experiment 2} This experiment tries the approach explored in the work by Geirhos et al. in which data augmentation with many distortion domains is used at training time to try to prepare the model to test well on a new distortion domain which was not seen during training. We train two models: one whose new unseen domain is the CT ring artifact, and another whose new unseen distortion domain is the rotation by 90°. We hypothesize that this should confirm what the Geirhos paper found regarding limitations of augmentation for generalization, but on a new dataset. In particular, the model should perform well on the 3 distortion types it has seen during training. However, like in Experiment 1, the model will not be able to generalize to the ring artifact or rotation which it has not seen before.

The OrganAMNIST dataset and its corresponding distortions are used once again. Both of the models are trained with all training instances with no distortion, and all training instances with uniform noise. One of the models additionally sees all training instances with rotation by 90°, whereas the other model instead additionally sees all training instances with the ring artifact. For each of the 2 models, we test the model on all 4 distortions’ test data.

This experiment uses the same hyperparameters as in experiment 1.

\paragraph{Experiment 3} This experiment adapts the technique from Experiment 1 with a domain adaptation architecture. With the new domain adaptation architecture, we allow the model to leverage the unlabeled target domain data that the previous experiments’ models were unable to leverage. Like in experiment 2, we train two models: one whose target domain is images with the ring artifact, and another whose target domain is images with rotation by 90°. We hypothesize that each model will perform well on both the source domain (no distortion) and its corresponding target domain test data, even though it never saw the labels for target domain data during training. We also expect no improved performance on the other two distortions, but are interested in seeing if domain adaptation provides sufficiently generalized features to improve performance on them even without an explicit approach aiming for them.

In this experiment, we switch to the DANN we describe earlier, which includes the domain classifier branch.

We once again use OrganAMNIST and the corresponding distortions. Both models are trained with all training instances with no distortion, which is used as the source domain. One model uses all training instances with ring artifacts as the unlabeled target domain; the other model instead uses all training instances with rotation by 90° as the unlabeled target domain. We again test each model on the test sets for each of the four distortions.

\begin{figure}[ht!]
  \includegraphics[width=0.9\linewidth]{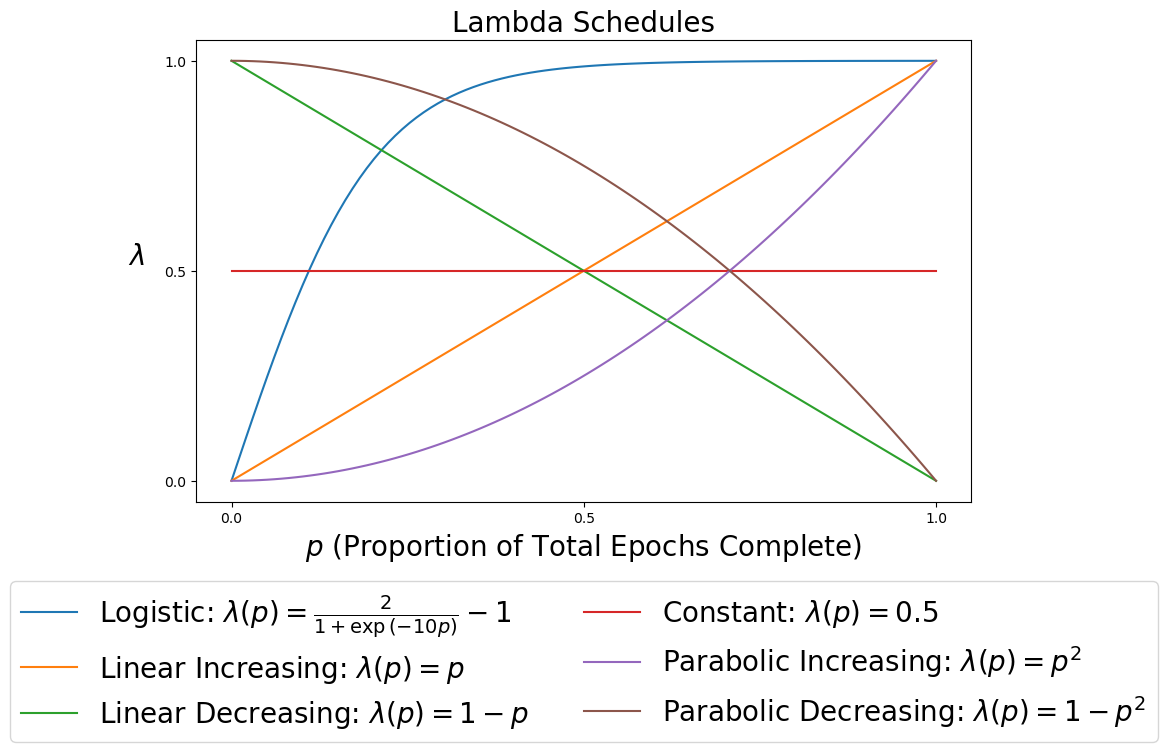}
  \caption{Explored schedules for hyperparameter $\lambda$. Both increasing and decreasing schedules are included to compare the effect of prioritizing adaptation later or earlier. Linear, parabolic, and logistic schedules are included to compare gradual shift in focus to a more frontloaded or backloaded focus on domain adaptation.}
  \label{figLambdaSchedules}
\end{figure}

\begin{table}[h!]
\caption{Validation Accuracy for Explored Lambda Schedules}
\begin{center}
\begin{tabular}{c|c}
\hline
$\lambda$ Schedule& \makecell{Validation Accuracy \\ $\left(\mathcal{T}=\text{Ring Artifact}\right)$}  \\ \hline
Logistic & 0.841 \\ \hline
Linear Inc. & 0.890 \\ \hline
Linear Dec. & 0.586 \\ \hline
Parabolic Inc. & \textbf{0.919} \\ \hline
Parabolic Dec. & 0.559 \\ \hline
Constant & 0.686 \\ \hline
\end{tabular}
\label{tableHyperparameterValidationAccuracy}
\end{center}
\end{table}

The hyperparameters are mostly the same as in experiment 1, with the addition of adaptation factor $\lambda$. In our initial runs of this experiment, we saw lower validation performance from our domain-adapted network compared to the baseline and augmentation-based training experiments. To address this, we explored several schedules for $\lambda$ in terms of proportion $p$ of total training epochs (which in this experiment is set to 50 epochs).

These schedules include the logistic schedule used by \cite{ganin2015unsupervised}, increasing \& decreasing linear schedules, increasing \& decreasing quadratic schedules, and a constant schedule, illustrated in Fig.~\ref{figLambdaSchedules}. We note that the increasing quadratic schedule performs best in our domain adaptation training experiment where the ring artifact is the target domain, as seen in Table \ref{tableHyperparameterValidationAccuracy} - the validation accuracy of 91.9\% observed with the parabolic increasing schedule was highest among all explored schedules. This suggests that an effective domain adaptation strategy is to focus primarily on the label prediction task for the early and middle epochs of training, and then increasingly shift the training's focus towards the domain adaptation task towards the end of training.

\section{Results \& Discussion}

\begin{figure}[ht!]
  \includegraphics[width=\linewidth]{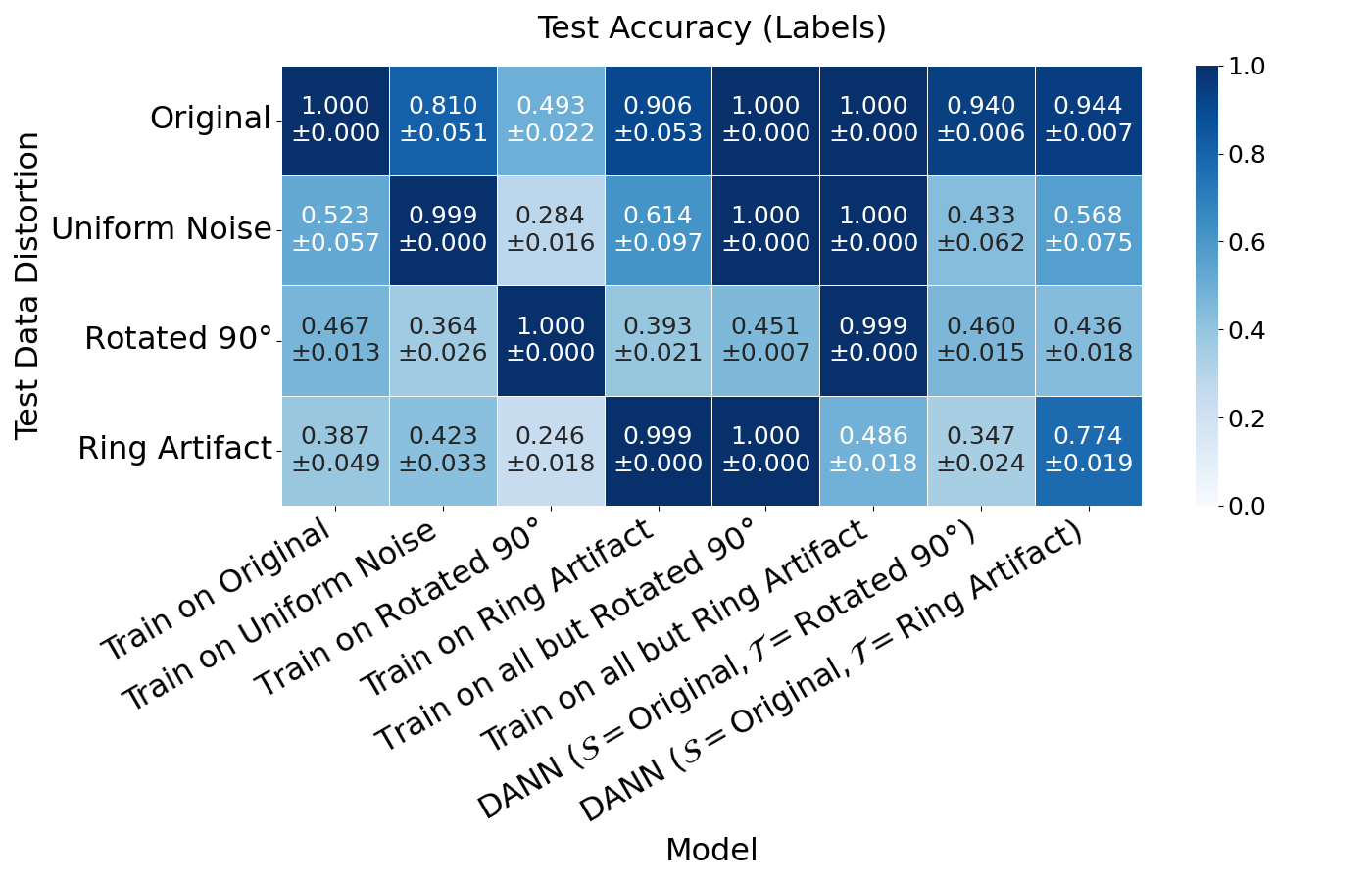}
  \caption{Classification accuracies for Experiments 1-3. Accuracies formatted as $\text{mean} \pm \text{std}$ over 5-fold cross validation. Given $C$=11 classes, random chance is at 1/11, or about 0.091. Columns 1-4 correspond to the baseline experiments with no adaptation strategy. Columns 5-6 correspond the approach of using traditional augmentation to prepare for an unseen domain. Columns 7-8 correspond to usage of a DANN.}
  \label{figExpt234ResultMatrix}
\end{figure}

We observe in the first 4 columns of Fig.~\ref{figExpt234ResultMatrix} that for each distortion we used in training, the base architecture learned that distortion well. As expected, the model generally does not generalize well to other distortions it did not train with. For example, the model trained on the original images (first column) achieved an accuracy of 100.0\% on average in 5-fold cross validation on test instances of original images (first row), but relatively low average accuracies on images with distortions (52.3\% on images with uniform noise in row 2, 46.7\% on rotated images in row 3, and 38.7\% on images with the ring artifact in row 4). Likewise, training on rotated images does not help with performance on the original images or on images with other distortions - although the model in the third column achieves average accuracy of 100.0\% on rotated images (third row), its average accuracy is low in the other domains (49.3\% in the first row, 28.4\% in the second, and 24.6\% in the fourth). Conversely, the models which were trained on a domain other than rotation show low performance in the third row on rotated data (average accuracy of 46.7\% in the first column, 36.4\% in the second, and 39.3\% in the fourth). Training on uniform noise or on the ring artifact seems to inform performance on the original images - in the second column, the model trained on uniform noise achieves near-perfect accuracy of 99.9\% on test images with uniform noise, and also an accuracy of 81.0\% on the original images, which is higher than its performance on the rotated and ring artifact images (third and fourth rows). Similarly, the model trained on the ring artifact achieves near-perfect accuracy of 99.9\% on test images with the ring artifact, and also a fairly high accuracy of 90.6\% on the original images.

Moving to Experiment 2, assessing augmentation as a generalization strategy, we see in the 5th and 6th columns of Fig.~\ref{figExpt234ResultMatrix} that again, for each domain we used in training, the base architecture learned that domain well.
We see that even though the model has seen a wider variation of domains, none will help with the domain omitted in training, as expected from the findings of \cite{geirhos2018generalisation}. In the 5th column, the model trained on all domains except the domain of rotated images achieved an accuracy of 100.0\% on the test data of domains it was exposed to (in the first, second, and fourth rows), but its accuracy of 45.1\% on the rotated data (third row) is not better than the accuracy of 46.7\% attained on the same data by the model which was only trained on original images (column 1). Likewise, in the 6th column, the model trained on all domains except the domain of images with the ring artifact achieved an accuracy of at least 99.9\% on the test data of domains it was exposed to (in the first, second, and third rows), but its accuracy of 48.6\% on the ring artifact data is not satisfactory.

The results we saw in Experiment 3 support our hypothesis that domain adaptation has potential to enable a classification model to perform well on medical images exhibiting a distortion which we lack labels for. Column 8 of Fig.~\ref{figExpt234ResultMatrix} in particular shows that the model which trained on undistorted data as the source domain and data with the ring artifact as the target domain retains a fairly high accuracy on the undistorted data (94.4\%). Importantly, it is able to classify the images with a ring artifact better than the baseline model of column 1. The domain-adapted model of column 8 shows an average accuracy of 77.4\% on ring artifact test data, whereas the baseline model in column 1 shows a much lower average accuracy of 38.7\% on the same test data. Furthermore, we can conclude that the domain adaptation strategy may have more potential than augmentation on data whose labels are not shown in training as column 6 also shows limited performance on ring artifact data (accuracy of 48.6\%). Unfortunately, the DANN which adapts to unlabeled ring artifact images is not able to perform on the ring artifact test set quite as well as a baseline model trained explicitly with labeled ring artifact images, which achieved a very high average accuracy of 99.9\%, suggesting that it is still preferable to learn directly from the labels if possible.

Although the domain adaptation strategy helped a model to adapt to ring artifact data, the same strategy was not as effective when applied to adapting to rotated data. In column 7 of Fig.~\ref{figExpt234ResultMatrix}, although the domain-adapted model retained a fairly high accuracy on its source domain of undistorted data (94.0\%), its accuracy of 46.0\% on its target domain of rotated data is no better than the baseline model (46.7\%). This result is somewhat expected, as convolutional neural networks do not have a natural mechanism for rotational invariance. Thus, domain adaptation is not necessarily appropriate for all target domains.

\begin{figure}[ht!]
  \includegraphics[width=\linewidth]{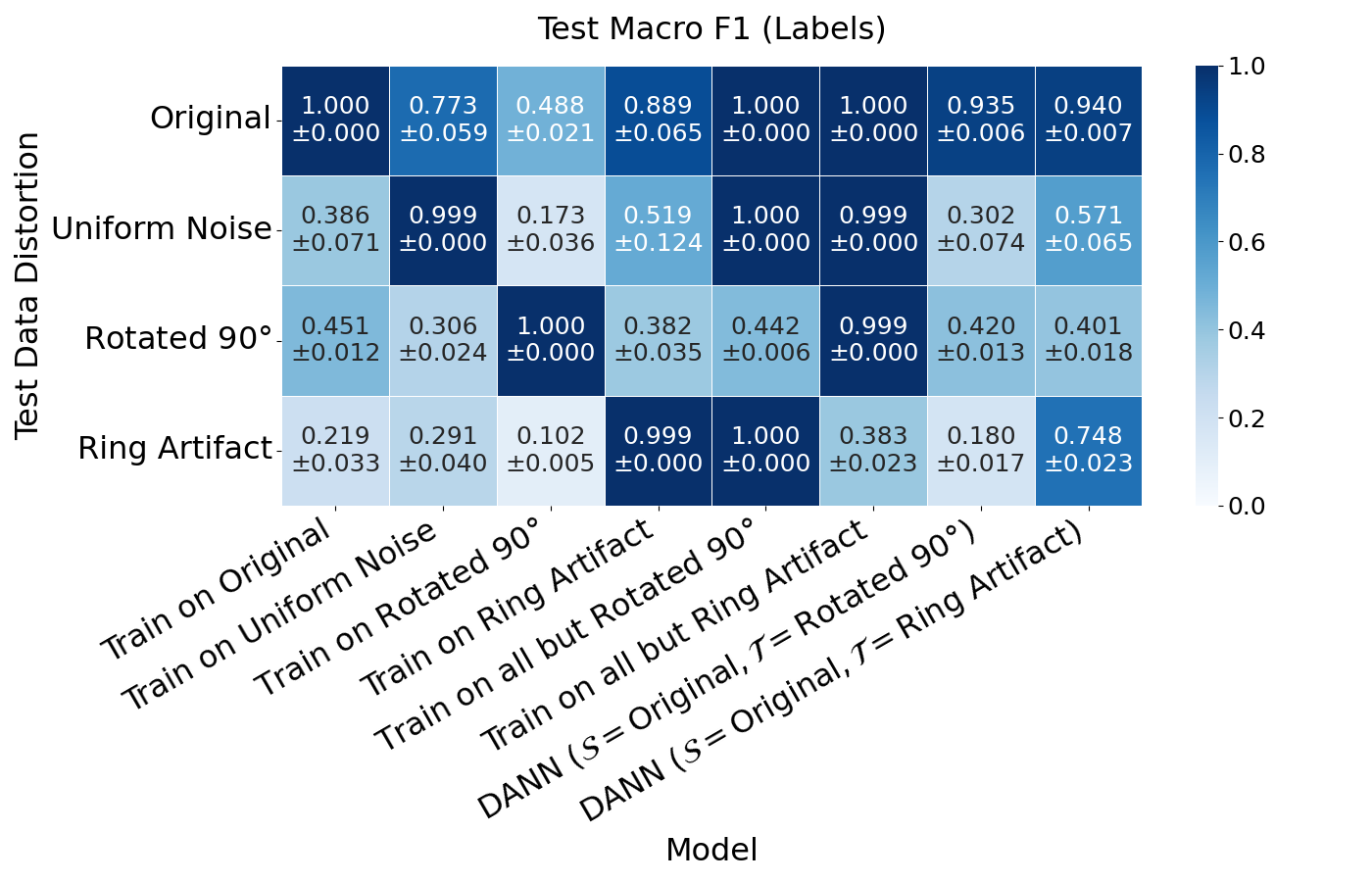}
  \caption{Macro-averaged F1 score for Experiments 1-3 in the same format as Fig.~\ref{figExpt234ResultMatrix}. F1 score broadly follows similar trends as accuracy.}
  \label{figExpt234ResultMatrixF1}
\end{figure}

\begin{figure}[ht!]
  \includegraphics[width=\linewidth]{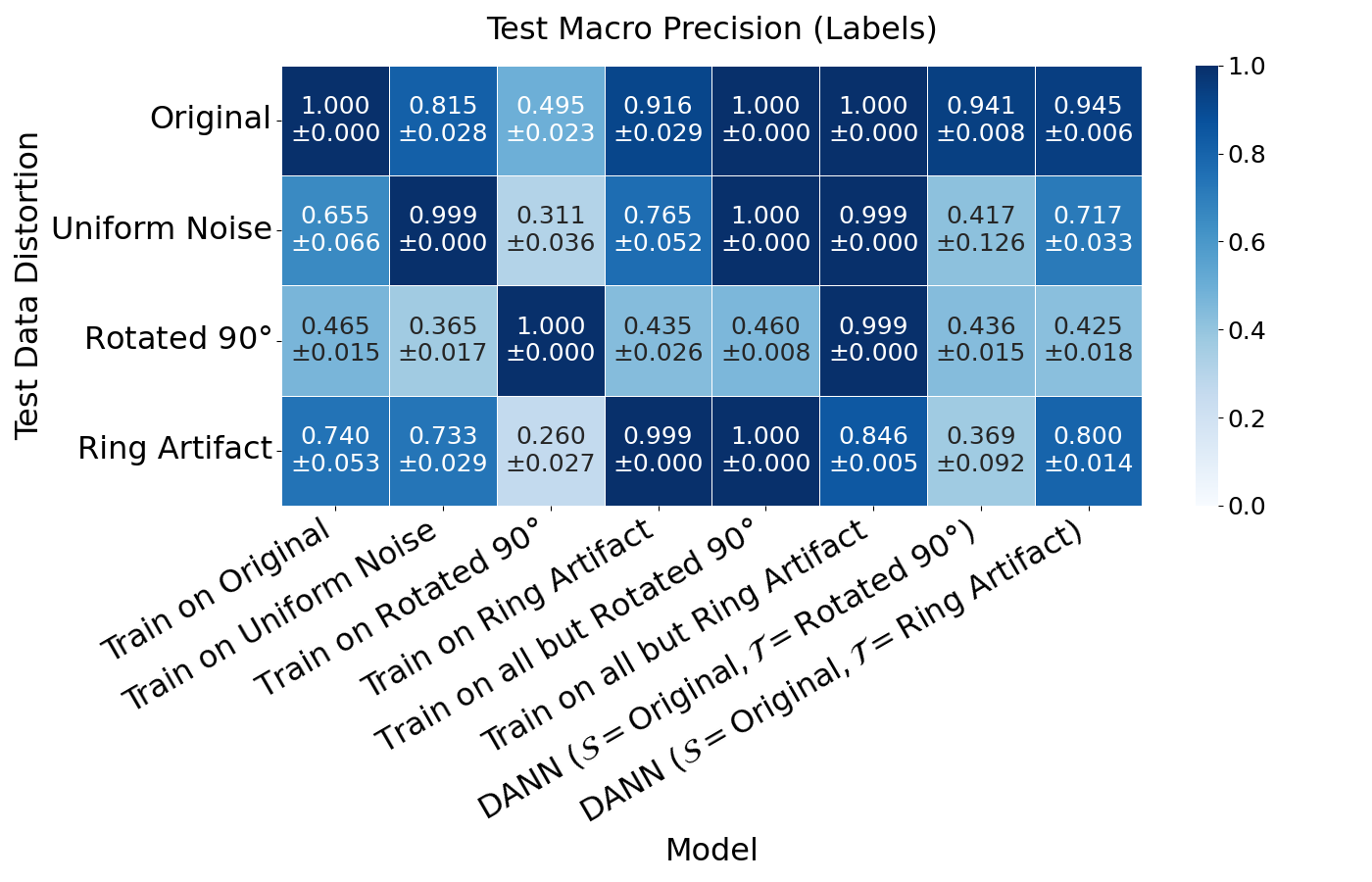}
  \caption{Macro-averaged precision for Experiments 1-3 in the same format as Fig.~\ref{figExpt234ResultMatrix}.}
  \label{figExpt234ResultMatrixPrecision}
\end{figure}

\begin{figure}[ht!]
  \includegraphics[width=\linewidth]{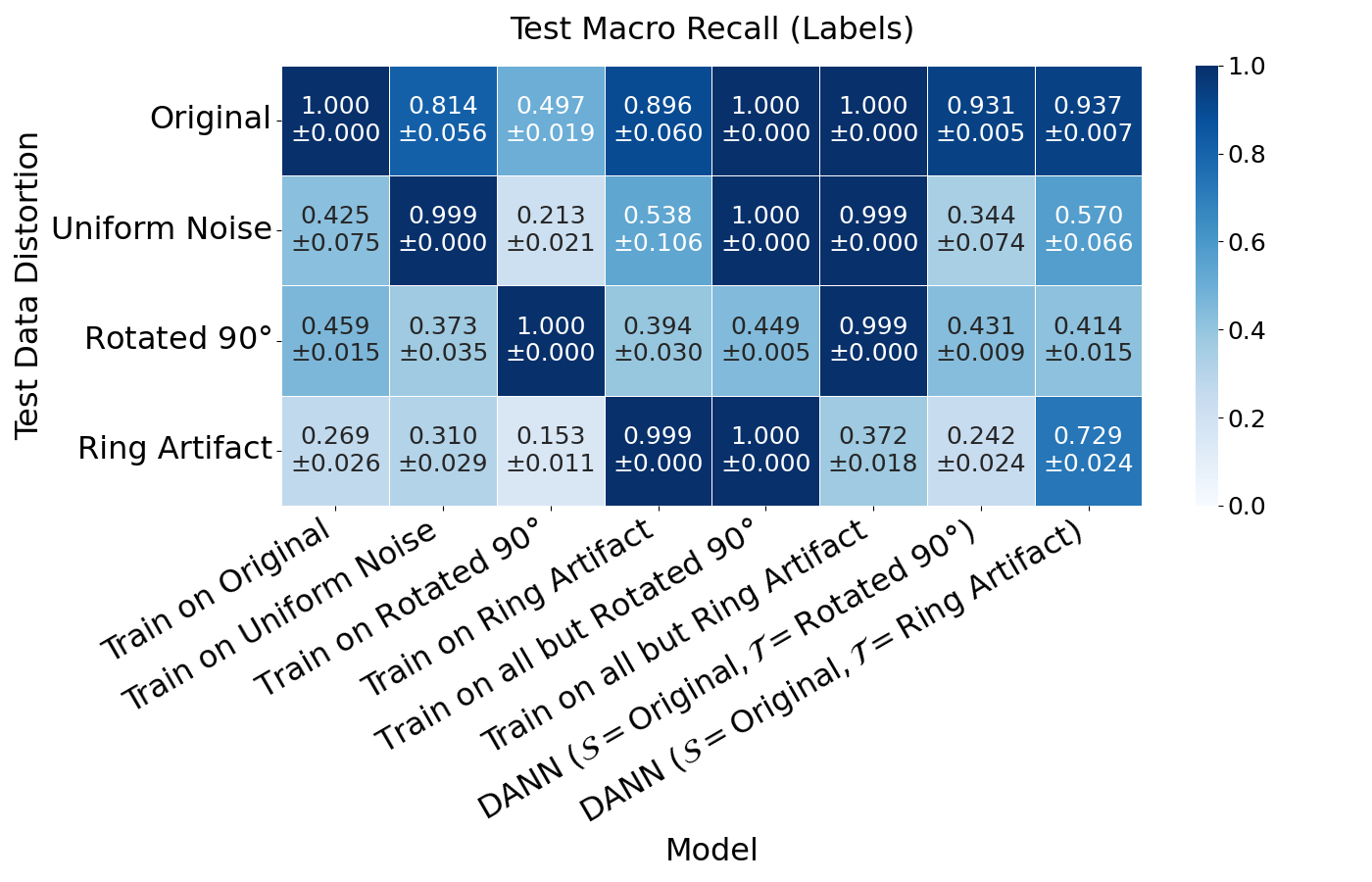}
  \caption{Macro-averaged recall for Experiments 1-3 in the same format as Fig.~\ref{figExpt234ResultMatrix}. F1 score broadly follows similar trends as accuracy.}
  \label{figExpt234ResultMatrixRecall}
\end{figure}

OrganAMNIST is an imbalanced classification problem. In the provided training set for example, the largest class (liver) has 6,164 samples, while the smallest class (right femur) has only 1,357 samples. Likewise, the validation set has 1,033 in its largest class and 225 in its smallest, while the test set has 3,285 in its largest and 784 in its smallest. As a result, we also collect the macro-averaged F1 (Fig.~\ref{figExpt234ResultMatrixF1}), precision (Fig.~\ref{figExpt234ResultMatrixPrecision}), and recall (Fig.~\ref{figExpt234ResultMatrixRecall}) for each experiment on OrganAMNIST data to help detect problems in predicting certain classes compared to others. F1, precision, and recall all broadly follow the same trends as accuracy. This suggests that for this problem, the class imbalance is not a major source of difficulty for the model.

\begin{figure}[ht!]
  \includegraphics[width=\linewidth]{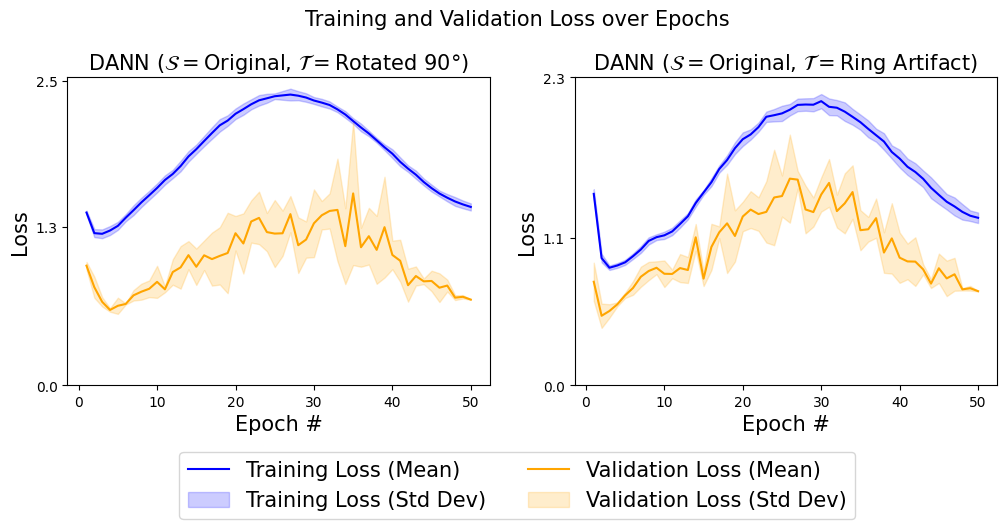}
  \caption{Training curves for the models from Experiment 3. For both models, training loss gradually rises, then gradually lowers. }
  \label{figLossCurves}
\end{figure}

We find the loss curves for the models of experiment 3 interesting (Fig.~\ref{figLossCurves}). For both models, training loss gradually rises at first, then gradually lowers. A potential explanation may be that as the domain classifier learns, the gradient reversal makes the features more difficult for the domain classifier to predict from, leading to an overall increase in loss. However, as the domain classifier gets better at using domain invariant features, and as the feature extractor gets better at providing useful domain invariant features for the label predictor, the loss comes down again. Although the validation loss fluctuates more, we still observe reasonable classification performance after training.

\section{Conclusion}

In this study, we provide empirical evidence that domain adaptation is a promising approach to directly address the challenge of adapting a medical imaging classification model to a new distribution of data without the expense common in medical imaging settings of obtaining labels for a new distribution. We also provide a technique for synthetic generation of distortions specific to CT to further test this domain adaptation approach to distortions which occur in CT scanners in practice, based in the principles of CT image acquisition.

Some exploration of training strategies could be considered to improve the overall classification accuracy of the domain adaptation approach towards the level we observed in models which were explicitly allowed to see both source and target domains (columns 5 or 6 of Fig.~\ref{figExpt234ResultMatrix}).

Our code implementations for this project, including exploration, preprocessing, training, and visualization scripts and Jupyter Notebooks, are available at: \\
\href{https://github.com/JustinCheung168/domain-adaptation-ct}{https://github.com/JustinCheung168/domain-adaptation-ct}.

While we achieved some success with domain adaptation techniques on OrganAMNIST, there are many more avenues to continue this research.

During this project we only were able to characterize a few types of distortions, with only the ring distortion being exclusive to CT imaging. There are several other CT imaging artifacts that can be encountered during CT imaging that could possibly benefit from domain adaptation techniques, such as metal artifacts, motion artifacts, or streak artifacts \cite{barrett2004artifacts}. 

Furthermore, due to time and compute constraints we only were able to look at the CT dataset OrganAMNIST. Additional datasets should be considered for future research to determine how effective domain adaptation is on different image datasets. Additionally, different imaging techniques, like magnetic resonance imaging (MRI) or ultrasound should also be considered.

It is important to note that our study is fundamentally limited by its usage of simulated artifacts without comparable real-world artifacts. Prior work has highlighted the impact of the gap between simulation and clinical reality on model performance \cite{bissoto2022artifact}. A reasonable next step for this work would be to acquire real clinical data exhibiting the artifacts of interest. Although the validation for such a study would require some labeling of this clinical data, the approach proposed here still allows utilization of a potentially larger body of unlabeled clinical data to contribute to artifact robustness. It is possible to frame closing the gap between simulated artifacts and clinical artifacts as a domain adaptation problem in itself - \cite{du2023deep} has successfully closed this gap in the case of adapting a metal artifact reduction network from simulated to real data.

We also do not extensively test the limits of domain adaptation with increasing intensity of our simulated ring artifact. It may be worthwhile for future work to determine whether adaptation begins to fail at a sufficiently high bound for gain error.

Our approach to shifting the isocenter relative to the image becomes computationally intractable as the shift from the center increases, which in conjunction with lacking accurate image position metadata makes simulating concentric artifacts for anatomies offset from the isocenter challenging in this approach.

Furthermore, we acknowledge that even the acquisition of unlabeled images with the artifact we seek to adapt to can be a challenge. We would ideally want the model to be robust even to new distributions seen only at test time, without needing to provide even unlabeled instances at training time. For this use case, approaches from the related field of domain generalization should be considered, though it should be noted factors such as the learning of domain invariant features with no target domain instances available at training time \cite{korevaar2023failure} as well as the characterization of out-of-distribution test sets \cite{bissoto2022artifact} remain as challenging problems to address for domain generalization.

\bibliographystyle{ACM-Reference-Format}
\bibliography{main}

\end{document}